\documentclass[11pt,a4paper]{article}
\usepackage[hyperref]{acl2020}
\usepackage{times}
\usepackage{latexsym}
\usepackage{adjustbox}

\usepackage{amsmath}
\usepackage{microtype}

\aclfinalcopy 

\title{``None of the Above":\\ Measure Uncertainty in Dialog Response Retrieval}
\author{Yulan Feng, Shikib Mehri, Maxine Eskenazi and *Tiancheng Zhao \\
  Language Technologies Institute, Carnegie Mellon University  \\
  *SOCO.AI \\
  \texttt{\{yulanf,amehri,max\}@cs.cmu.edu} \\
  *\texttt{tianchez@soco.ai}}

\date{}

\begin{document}
\maketitle

\begin{abstract}
This paper discusses the importance of uncovering uncertainty in end-to-end dialog tasks and presents our experimental results on uncertainty classification on the processed Ubuntu Dialog Corpus\footnote{Our datasets for the NOTA task are released at https://github.com/yfeng21/nota\_prediction}. We show that instead of retraining models for this specific purpose, we can capture the original retrieval model's underlying confidence concerning the best prediction using trivial additional computation. 
\end{abstract}

\section{Introduction}
Uncertainty modeling is a widely explored problem in dialog research. Stochastic models like deep Q-networks \cite{tegho2017uncertainty}, Gaussian processes \cite{6601004}, and partially observable Markov decision process \cite{Roy:2000:SDM:1075218.1075231} are often used in spoken dialog systems to optimize dialog management by explicitly estimating uncertainty in policy assignments.  

However, these approaches are either computationally intensive \cite{gal2015dropout} or require significant work on refining policy representations \cite{6601004}. Moreover, most current uncertainty studies in dialog focus on the dialog management component. End-to-end (E2E) dialog retrieval models jointly encode a dialog and a candidate response \citep{wu2016sequential,zhou2018multi}, assuming the ground truth is always present in the candidate set, which is not the case in production. \citet{larson2019evaluation} recently showed that classifiers that perform well on in-scope intent classification for task-oriented dialog systems struggle to identify out-of-scope queries. The response selection task in the most recent Dialog System Technology Challenge \cite{dstc19task1} also explicitly mentions that ``none of the proposed utterances is a good candidate" should be a valid option. 

The goal of this paper is to set a new direction for future task-oriented dialog system research: while retrieving the best candidate is crucial, it should be equally important to identify when the correct response (i.e. ground truth) is not present in the candidate set. In this paper, we measure the E2E retrieval model's capability to capture uncertainty by inserting an additional ``none of the above" (NOTA) candidate into the proposed response set at inference time. 


The contributions of this paper include: (1) demonstrating that it is crucial to learn the relationship amongst the candidates as a set instead of looking at point-wise matching to solve the NOTA detection task. As a result, the logistic regression (LogReg) approach proposed here consistently achieves the best performance compared to several strong baselines. (2) extensive experiments show that the raw output score (\emph{logits}) is more informative in terms of representing model confidence than normalized probabilities after the Softmax layer.

\section{Related Work}
Our use of NOTA to measure uncertainty in dialog response is motivated by the design of student performance assessment in psychology studies.  

Test creators often include NOTA candidates in multiple-choice design questions, both as correct answers and as distractors. How the use of NOTA affects the difficulty and discrimination of a question has been discussed widely \cite{guidlines,psychologycls}. For assessment purposes, a common finding is that using NOTA as the correct response increases question difficulty, and also lures high- and low-performing students toward distractors \cite{psychologycls}.

Returning a NOTA-like response is a common practice in dialog production systems \cite{watson}. The idea of adding the NOTA option to a candidate set is also widely used in other language technology fields like speaker verification \cite{6289356}. However, the effect of adding NOTA is rarely introduced in dialog retrieval research problems. To the best of our knowledge, we are the first to scientifically evaluate a variety of conventional approaches for retrieving NOTA in the dialog field.



\section{Methods}
\subsection{Ubuntu Dataset}
All of the experiments herein use the Ubuntu \cite{lowe-etal-2015-ubuntu} Dialog Corpus, which contains multi-turn, goal-oriented chat logs on the Ubuntu forum. For next utterance retrieval purposes, we use the training data version that was preprocessed by \citet{mehri2019multigranularity}, where all negative training samples (500,127) were removed, and, for each context, 9 distractor responses were randomly chosen from the dataset to form the candidate response set, together with the ground truth response. For the uncertainty task, we use a special token \emph{\_NOTA} to represent the ``none of the above" choice, as in multiple-choice questions. More details on this NOTA setup can be found in Sections \ref{ssec:direct} and \ref{ssec:threshold}.
The modified training dataset has 499,873 dialog contexts, and each has 10 candidate responses. The validation and test sets remain unchanged, with 19,561 validation samples and 18,921 test samples.


\subsection{Dual LSTM Encoder}
\label{ssec:lstm}
The LSTM dual encoder model consists of two
single-layer, uni-directional encoders, one to encode the embedding ($c$) of the context and one to encode the embedding ($r$) of
the response. The output function is computed as a dot product of the two, $f(r,c) = c^Tr$. This model architecture has already been shown to perform well for the Ubuntu dataset \cite{lowe-etal-2015-ubuntu,kadlec2015improved}. We carry out experiments with the following variants of the vanilla model for training:

\paragraph{Binary}
This is the most common training method for next utterance ranking on the Ubuntu corpus. With training data prepared in the format of [CONTEXT] [RESPONSE] [LABEL], the model performs binary classification on each sample, predicting whether a given response is the ground truth. The binary cross-entropy between the label and $\sigma(f(r,c))$ following a sigmoid layer is used as the loss function.

\paragraph{Selection}
As the validation and test datasets are both in the format of \emph{[CONTEXT] [RESPONSE]*x}, where x is usually 10, we train the selection model in the same way. For this model, following a softmax layer, the loss is calculated by the negative log likelihood function: 

\begin{equation}
\mathcal{L} = -\log \left(\frac{exp(f(r_{\text{ground truth}},c)}{\sum_{i=1}^x exp(f(r_{i},c))}\right)
\end{equation}

\paragraph{Dropout}
\citet{gal2015dropout} found that dropout layers can be used in neural networks as a Bayesian approximation to the Gaussian process, and thus have the ability to represent model uncertainty in deep learning. Inspired by this work, we add a dropout layer after each encoder's hidden layer at training time. At inference, we have the dropout layer activated and pass each sample through $n$ times, and then make the final prediction by taking a majority vote among the $n$ predictions. Unlike the other models, the NOTA binary classification decision is not based on the output score itself, but rather is calculated on the score variance of each response. 

\subsection{Experimental Setup}
\label{ssec:setup}
\paragraph{LSTM}
For the LSTM models, unless otherwise specified, the word embeddings are initialized randomly with a dimension of 300, and a hidden size of 512. The vocabulary is constructed of the 10000 most common words in the training dataset, plus the \emph{\_UNK} and \emph{\_PAD} special tokens. We use the Adam algorithm \cite{adam} for optimization with a learning rate of 0.005. The gradients are clipped to 5.0. With a batch size of 128, we train the model for 20 epochs, and select the best checkout based on its performance on the validation set. In the dropout model, we use a dropout probability of $50\%$. 

\paragraph{LogReg}
For the logistic regression model, we train on the validation set's LSTM outputs with the same hyperparameter (where applicable to LogReg) setup as in the corresponding LSTM model.
\section{Experiments}

\subsection{Direct Prediction}
\label{ssec:direct}
For the direct prediction experiment, we randomly choose $50\%$ of the response sets and replace the ground truth responses with the \emph{\_NOTA} special token (we label this subset as \emph{isNOTA}). For the other $50\%$ samples, we replace the first distractor with the  \emph{\_NOTA} token (we label this subset as \emph{notNOTA}). By using this setup, we ensure that a \emph{\_NOTA} token is always present in the candidate set. Although making decisions based on logits (\emph{Directlogits}) or probability (\emph{DirectProb)} yields the same argmax prediction, we collect both output scores for the following LogReg model (details in Section \ref{ssec:logreg}). Concretely, the final output $y'$ of a direct prediction model is: 
\begin{equation}
    y' = \text{argmax}_{r \in A \bigcup \{\text{\_NOTA}\}} f(r, c)
\end{equation}

\subsection{Threshold}
\label{ssec:threshold}
Another common approach toward returning NOTA is to reject a candidate utterance based on confidence score thresholds. Therefore, in the threshold experiments, with the same preprocessed data as in Section \ref{ssec:direct}, we remove all \emph{\_NOTA} tokens at the inference model's batch preparation stage, leaving 9 candidates, thus $50\%$ of the response sets (the \emph{isNOTA} set) with no ground truth present. After the model outputs scores for each candidate response, with the predefined threshold, it further decides whether to accept the prediction with the highest score as its final response, or to reject the prediction and give NOTA instead. We investigate the performance of setting the threshold based on probability (\emph{ThresholdProb}) and logits (\emph{ThresholdLogits}) respectively. Concretely, the final output $y'$ is given by: 
\begin{equation}
y' =\begin{cases}
    \text{\_NOTA if } f(r, c) < \text{threshold}\\
    \text{argmax}_{r \in A} f(r, c)
    \end{cases}
\end{equation}

\subsection{Logistic Regression}
\label{ssec:logreg}
 We feed the output scores of the LSTM models for all candidate answers as input features to the LogReg model consisting of a single linear layer and a logistic output layer. Separate LogReg models are trained for different numbers of candidates. The probability output indicates whether the previous model's prediction is ground truth or just the best-scoring distractor. Since LogReg can see output scores from all candidate responses, it is trained to model the relationship amongst all the candidates, making it categorically different from the binary estimation mentioned in Section~\ref{ssec:direct} and \ref{ssec:threshold}. Note that at inference time, LogReg works essentially as a threshold method. The final output is determined by: 
 \begin{equation}
y' =\begin{cases}
    \text{\_NOTA if } \text{LogReg}(\{f(r_i, c)\}) < 0.5\\
    \text{argmax}_{r \in A} f(r, c)
    \end{cases}
\end{equation}

where input to the LogReg model $f(r_i,c)$ is the output of LSTM models, either in logits or normalized form, as previously defined in subsection \ref{ssec:lstm}.

\subsection{Metric Design}
\label{ssec:metric}
Dialog retrieval tasks often use recall out of k ($R_x@k$) as a key metric, measuring out of $x$ candidates how often the answer is in top-k. In this paper, we focus on the top-1 accuracy $R_x@1$ ($R_x$ for short) with a candidate set size of $x$, where $x\in \{2,5,10,20,40,60,80,100\}$. The recall metric is modified for uncertainty measurement purposes, and is further extended to calculate the NOTA accuracy out of $x$ ($N_x$), and F1 scores for each class ($NF1_x,GF1_x$). Let $D=\{c, y\}$ and $D_{n}=\{c, \text{\textit{isNOTA}}\}$ be the two subparts of data that correspond to samples that are \textit{notNOTA} and \textit{isNOTA} respectively, the above metrics are computed by:

\begin{align}
 R_x &= \frac{\sum_{y \in D} (y' = y)}{|D|} \\
 N_x &= \frac{\sum_{y \in D_n} (y' = y) + \sum_{y \in D} (y' \ne \text{\_NOTA})}{|D|+|D_n|} 
\end{align}

In Equation (6), the numerator represents correctly predicted (same as in Equation (5)) plus other true negative \textit{isNOTA} predictions, where the model correctly predicts \textit{notNOTA}, but fails to choose the ground truth.

The positive class in $NF1_x$ is the \emph{isNOTA} class, and the positive class in $GF1_x$ is the \emph{notNOTA} class. 

\subsection{More Candidates}
 In real-world problems, retrieval response sets usually have many more than 10 candidates. Therefore, we further test the selection and binary models on a bigger reconstructed test set. For each context, we randomly select 90 more distractors from other samples' candidate responses, producing a candidate response set of size 100 for each context.

\section{Results and Analysis}

\begin{table}[!b]
\centering
\begin{adjustbox}{width=0.5\textwidth}
\begin{tabular}{rll|lllll}
\cline{1-7}
\multicolumn{1}{l|}{}                                             & $R_{10}$ & $N_{10}$  & $NF1_{10}$ & $GF1_{10}$ & Average F1 &  \\\cline{2-7}
\multicolumn{1}{l|}{Selection Model (original data)} &         &       &         &         &             &  \\
\multicolumn{1}{r|}{Direct Predict}                               & 56.12 & 61.48 & 52.82 & 67.46 &60.14&  \\
\multicolumn{1}{r|}{+LogReg (Logits)}        & 55.98 & 87.81 & 86.96 & 88.56 &\textbf{87.76}&  \\
\multicolumn{1}{r|}{+LogReg (Softmax)}       &50.94 & 74.30 & 74.46 & 74.15 &74.31&  \\
\multicolumn{1}{r|}{Logits Threshold (=0.5)}                       & 50.10 & 64.28 & 62.84 & 65.61 &64.22& \\
\multicolumn{1}{r|}{+LogReg}                  & 62.81 & 80.45 & 80.49 & 80.42 &80.45&  \\
\multicolumn{1}{r|}{Softmax Threshold (=0.55)}                    & 48.76 & 60.10 & 59.69 & 60.50 &60.09&  \\
\multicolumn{1}{r|}{+LogReg}                  &63.64 & 78.50 & 80.17 & 76.52 &78.34&  \\ \cline{1-7}

\multicolumn{1}{l|}{Selection Model (\emph{\_NOTA})}          &         &       &         &         &             &  \\
\multicolumn{1}{r|}{Direct Predict}                               & 55.43 & 63.07 & 54.28 & 69.03 &61.66&  \\
\multicolumn{1}{r|}{+LogReg (Logits)}        & 40.66 & 78.19 & 78.80 & 77.53 &78.16&  \\
\multicolumn{1}{r|}{+LogReg (Softmax)}       &51.63 & 77.94 & 78.21 & 77.67 &77.94&  \\
\multicolumn{1}{r|}{Logits Threshold (=2.0)}                              & 48.44 & 61.32 & 57.75 & 64.32 &61.03&  \\
\multicolumn{1}{r|}{+LogReg}                  &60.73 & 79.22 & 79.11 & 79.33 &\textbf{79.22}&  \\
\multicolumn{1}{r|}{Softmax Thtrshold (=0.5)}                            & 48.18 & 59.06 & 57.32 & 60.67 &59.00&  \\
\multicolumn{1}{r|}{+LogReg}                  & 61.08 & 78.01 & 79.75 & 75.94 &77.84&  \\\cline{1-7}

\multicolumn{1}{l|}{Binary Model}          &         &       &         &         &             &  \\
\multicolumn{1}{r|}{Direct Predict}                               & 35.73 & 61.72 & 63.54 & 59.72 &61.63&  \\
\multicolumn{1}{r|}{+LogReg (Logits)}        &35.64 & 94.08 & 93.72 & 94.40 &\textbf{94.06}&  \\
\multicolumn{1}{r|}{+LogReg (Softmax)}       &25.42 & 85.06 & 85.41 & 84.69 &85.05&  \\
\multicolumn{1}{r|}{Logits Threshold (=1.0)}                              & 41.64 & 61.50 & 57.77 & 64.62 &61.20&  \\
\multicolumn{1}{r|}{+LogReg }                  & 51.58 & 77.15 & 76.74 & 77.55 &77.14&  \\
\multicolumn{1}{r|}{Softmax Threshold (=0.4)}                            & 39.70 & 54.96 & 51.83 & 57.70 &54.77&  \\
\multicolumn{1}{r|}{+LogReg}                  &52.00 & 74.40 & 76.43 & 71.99 &74.21&  \\ \cline{1-7}

\multicolumn{1}{l|}{Dropout Model}          &         &       &         &         &             &  \\
\multicolumn{1}{r|}{Direct Predict}                               & 28.57 & 50.13 & 1.48 & 66.61 &34.05&  \\
\multicolumn{1}{r|}{+LogReg (Logits)}        & 19.21 & 66.89 & 61.87&70.74 &\textbf{66.30}&\\
\multicolumn{1}{r|}{+LogReg (Softmax)}       & 21.73 & 50.49 & 56.37 & 42.79 &49.58&\\
\multicolumn{1}{r|}{Logits Variance Threshold (=0.1)}                              & 13.73  & 51.89 & 57.15  & 45.15   &51.15&  \\
\multicolumn{1}{r|}{+LogReg} & 20.87 & 56.13 & 40.18 & 65.37 &52.78&\\
\multicolumn{1}{r|}{Softmax Variance Threshold (=0.001)}  &   22.22      &     50.03  &  38.98 &    57.69&48.33&  \\
\multicolumn{1}{r|}{+LogReg }  & 23.84 & 57.21 & 60.87 & 52.81 &56.84&  \\

\end{tabular}

\end{adjustbox}
\caption{\label{results-table}Results on 10 candidates. $R$ represents recall, $N$ represents binary NOTA classification accuracy, $N F1$ represents the F1 score on the NOTA class, and $G F1$ represents the F1 score on the ground-truth-present class. Average F1 is the average of $N F1$ and $G F1$.}

\end{table}


Table \ref{results-table} summarizes the experimental results. Due to space limitation, this table only displays results on 10 candidates. Complete results on other numbers of candidates, which have similar performance patterns as 10, are found in the Appendix. The thresholds and hyperparameters are tuned on the validation set according to the highest average F1 score. For the selection model, in addition to the original dataset, we also train the model on a modified training dataset, containing \emph{\_NOTA} choices as in inference datasets, with the same set of hyperparameters. As expected, since there are now fewer real distractor responses, training including \emph{\_NOTA} improves the model's NOTA classification performance, but sacrifices recall scores, which is not desirable. In all the models, regardless of the training dataset used and the model architecture, adding a logistic regression on top of the LSTM output significantly improves average F1 scores. Specifically, the highest F1 scores are always achieved with logits scores as LogReg input features. These results show that, though setting a threshold is a common heuristic to balance true and false acceptance rates \cite{larson2019evaluation}, its NOTA prediction performance is not comparable to the LogReg approach, even after an exhaustive grid-search of best thresholds. This finding is underlined by receiver operating characteristic (ROC) curves on the validation set

\begin{figure}[!h]
    \centering
        \includegraphics[width=0.45\textwidth]{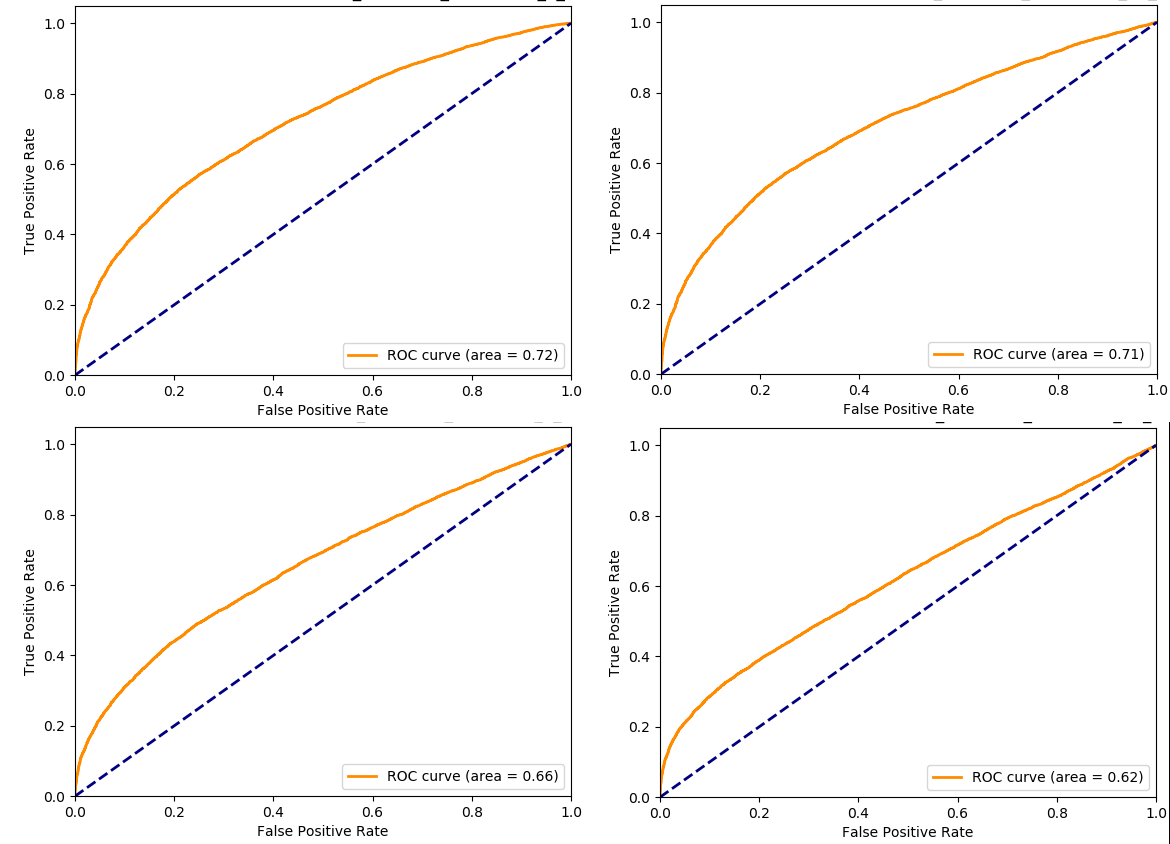} 
        \caption{\label{roc-fig}Merged ROC curves for LSTM outputs with the original selection model. Top left, top right, bottom left, and bottom right represent plots for \emph{ThresholdLogits},\emph{Directlogits}, \emph{ThresholdProb}, and \emph{DirectProb} respectively}
\end{figure}

\begin{figure}[!h]
    \centering
        \includegraphics[width=0.45\textwidth]{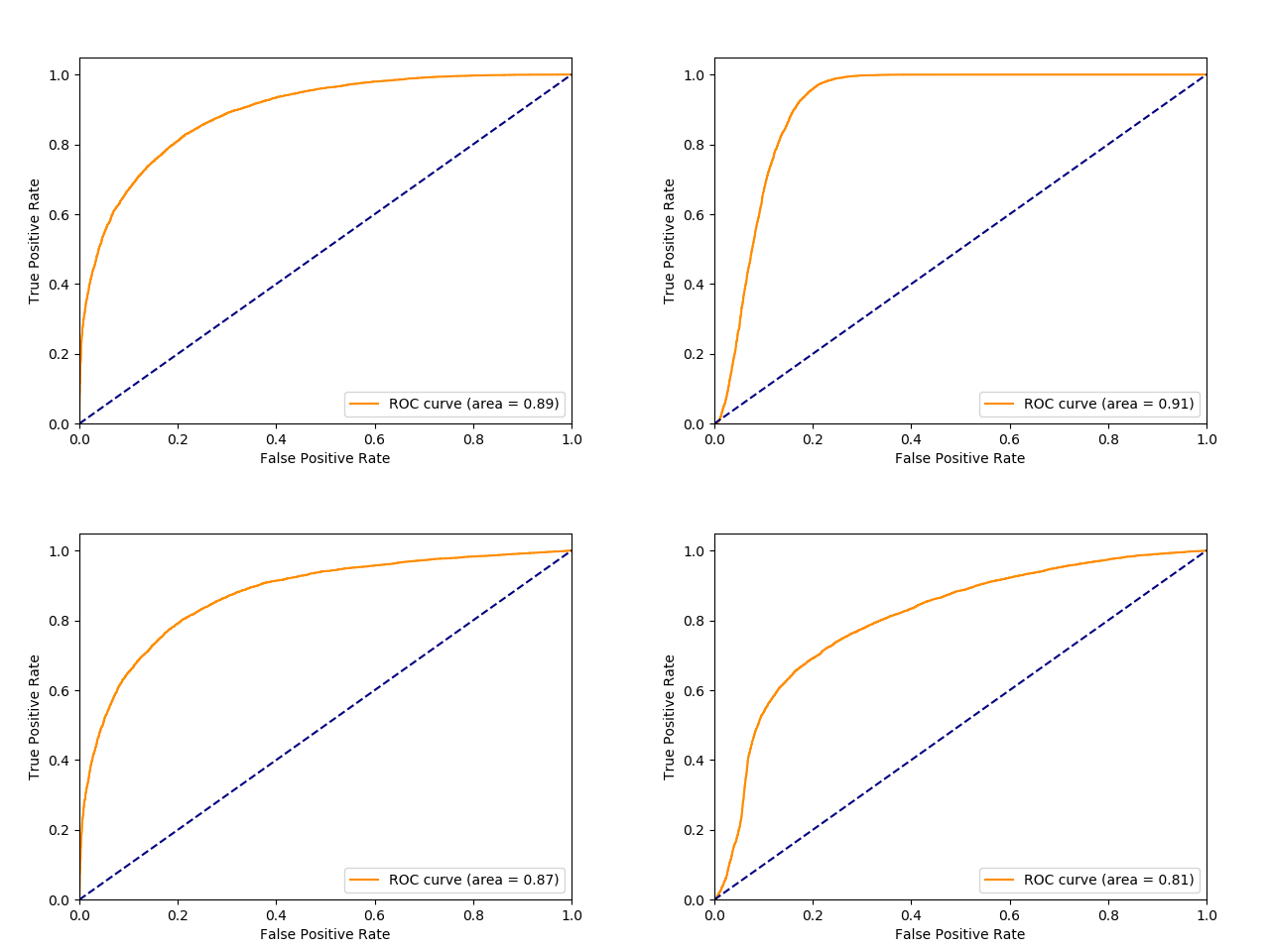} 
        \caption{\label{roc-logreg-fig}ROC curves for LogReg outputs with the original selection model's output logits as input features. Top left, top right, bottom left, and bottom right represent plots for \emph{ThresholdLogits},\emph{Directlogits}, \emph{ThresholdProb}, and \emph{DirectProb} respectively}
\end{figure}

Figure \ref{roc-fig} shows the ROC curves for predicting NOTA directly with LSTM. Figure \ref{roc-logreg-fig} shows ROC plots for predicting NOTA with LogReg in the same order as Figure \ref{roc-fig}, where a separate LogReg model is trained for each score setting. In both figures, the areas under curve (AUC) indicate that logits serves as a more discriminative confidence score compared to the normalized softmax score. Comparing the top right plots in both Figures, we can see that with the same set of logits scores as threshold criteria, AUC is boosted from 0.71 to 0.91 with the additional LogReg model, providing further evidence that LogReg significantly outperforms the LSTM models in this NOTA classification task. 

\begin{figure}[htb]
    \centering
        \includegraphics[width=0.5\textwidth]{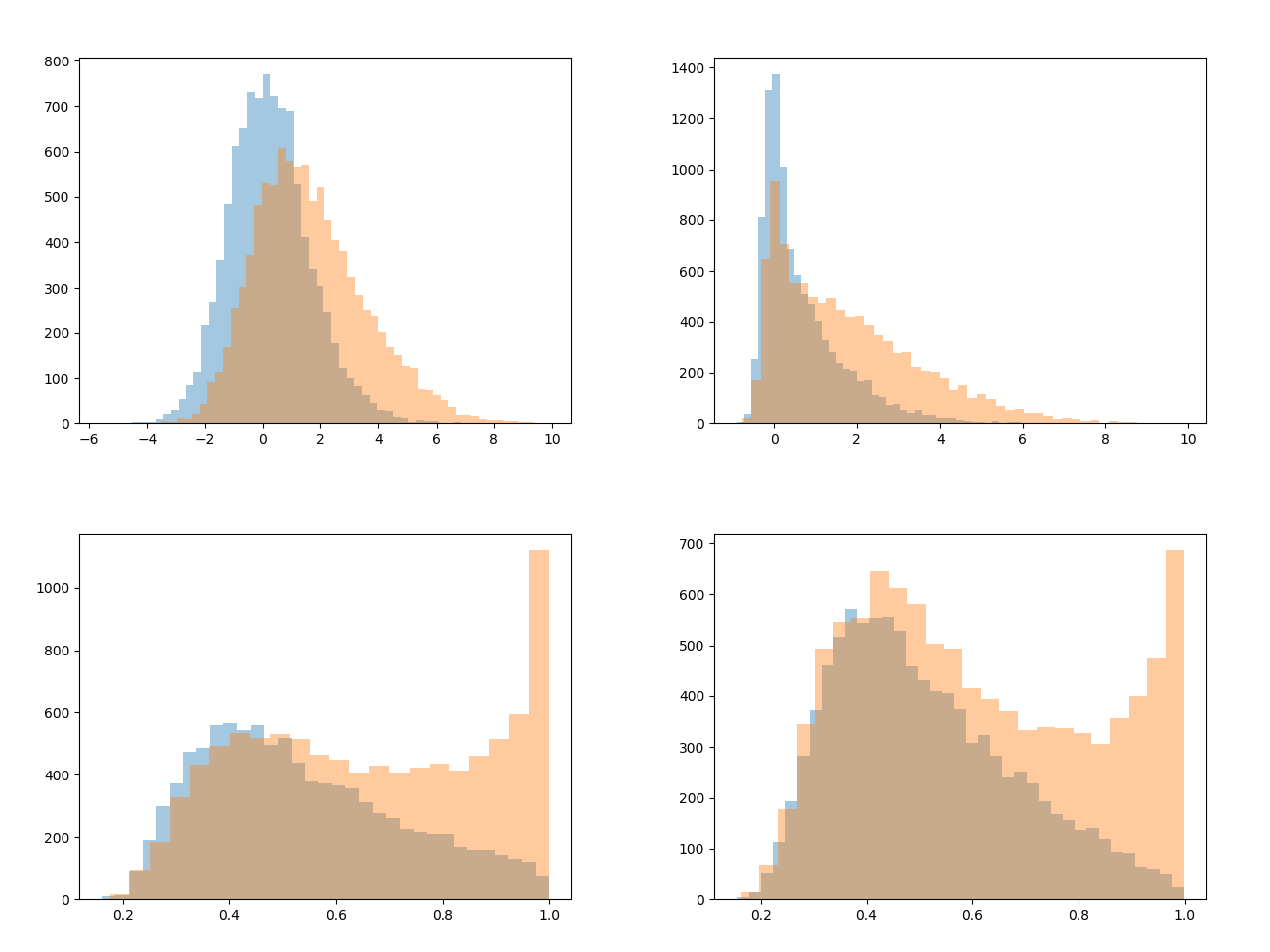} 
        \caption{\label{distribution-fig}Distribution of max scores as predicted by the original selection model, with scores (logits or probability) on the x-axis, and number of samples on the y-axis. Blue plot represents the \emph{isNOTA} subset, and orange plot represents the \emph{notNOTA}. Top left, top right, bottom left, and bottom right represent plots for \emph{ThresholdLogits},\emph{Directlogits}, \emph{ThresholdProb}, and \emph{DirectProb} respectively}
\end{figure}

With the selection model trained on the original dataset, Figure \ref{distribution-fig} shows the model's distribution of max scores on the validation set. We see that there are apparent differences between \emph{isNOTA}' and \emph{notNOTA}'s best score distributions. This is an encouraging observation because it suggests that current retrieval models can already distinguish good versus wrong responses to some extent. Note that as the \emph{\_NOTA} token is not included in training, for direct prediction tasks, the \emph{\_NOTA} token is encoded as an \emph{\_UNK} token at inference time. The tails of the \emph{isNOTA} plot in both the \emph{DirectLogits} and \emph{DirectProb} graphs suggest that the model will, very rarely, pick the unknown token as the best response. 
 
\begin{figure}[h!]
    \centering
        \includegraphics[width=0.5\textwidth]{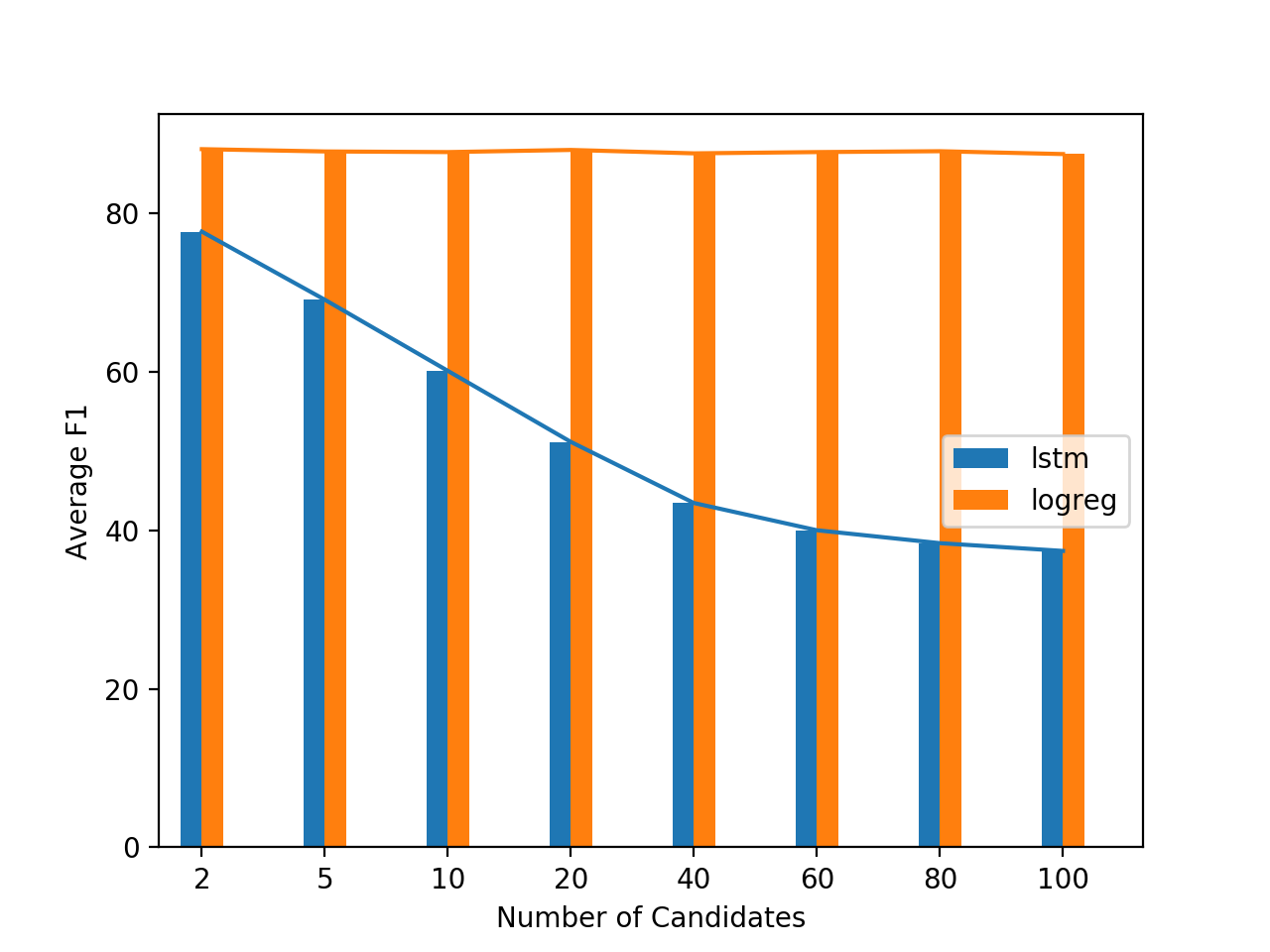} 
        \caption{\label{trend-fig}Average F1 scores with different numbers of response candidates, where the LSTM model stays the same, and LogReg is separately trained for each number setting. The left blue bars represent LSTM direct prediction, and the right orange bars represent LogReg results with logits input. }
\end{figure}

Figure \ref{trend-fig} shows the average F1 score trends with the original selection model on the test set with 100 distractors. The plot shows the trend that with more distractors, the LSTM model struggles to determine the presence of ground truth, while the LogReg model performs consistently well. The complete results of this extended test set are in the Appendix. 


\section{Discussion}
With \textit{\_NOTA} options in the training data, the models learn to sometimes predict \textit{\_NOTA} as the best response, resulting in more false-positive \textit{isNOTA} predictions at inference time. Also, by replacing various ground truths and strong distractors with NOTA, the model has fewer samples to help it learn to distinguish between different ground truths and strong distractors/ Thus it performs less well on borderline predictions (scores close to the threshold). This behavior results in some selection methods trained on the dataset containing \textit{\_NOTA} tokens performing worse than when they are trained on the original dataset. This motivates us to advocate the proposed LogReg approach instead of the conventional “add a NOTA choice” method. 

Another prominent advantage of the LogReg approach is that it does not require data- or model-dependent input like embedding vectors or hidden layer output. Instead, it takes logits or normalized scores, both of which can be output from any models. This feature makes our approach insensitive to the underlying architecture.


\section{Conclusions}
We have created a new NOTA task on the Ubuntu Dialog Corpus, and have proposed to solve the problem by learning the response set representation with a binary classification model. We hope the dataset we release will be used to benchmark future dialog system uncertainty research.


\bibliography{acl2020}
\bibliographystyle{acl_natbib}

\clearpage

\appendix
\section{Appendices}

\label{sec: appendix}

\subsection{More Plots}

\subsection{Complete Results}

\begin{table}[!h]
\centering
\begin{adjustbox}{width=0.5\textwidth}
\begin{tabular}{lll|lll}
\cline{1-6}
\multicolumn{6}{c}{50\% NOTA Test Results On More Distractors (\%)}\\ \cline{1-6}
\#Candidates & R & N  & N F1 & G F1 & Average F1\\\cline{1-6}
Direct Predict      &  &  &  &  &\\                    
2
 & 66.77 & 78.00 & 80.22 & 75.21 & 77.72\\
5
 & 62.14 & 69.17 & 67.86 & 70.38 & 69.12\\
10
 & 56.04 & 61.48 & 52.82 & 67.46 & 60.14\\
20
 & 48.09 & 55.81 & 36.11 & 66.22 & 51.17\\
40
 & 39.79 & 52.46 & 20.90 & 66.02 & 43.46\\
60
 & 34.96 & 51.20 & 14.12 & 65.92 & 40.02\\
80
 & 31.50 & 50.84 & 10.70 & 66.09 & 38.39 \\
 100
 & 29.10 & 50.59 & 8.69 & 66.13 & 37.41\\\cline{1-6}

+LogReg          &  &  &  &  &\\
2
 & 66.72 & 88.19 & 87.26 & 88.99 & 88.13\\
5
 & 62.07 & 87.90 & 87.01 & 88.67 & 87.84\\
10
 & 55.98 & 87.81 & 86.96 & 88.56 & 87.76\\
20
 & 48.07 & 88.08 & 87.27 & 88.79 & 88.03\\
40
 & 39.78 & 87.64 & 86.89 & 88.30 & 87.60\\
60
 & 34.95 & 87.80 & 87.07 & 88.46 & 87.76\\
80
 & 31.49 & 87.92 & 87.11 & 88.63 & 87.87\\
100
 & 29.10 & 87.55 & 86.84 & 88.18 & 87.51\\
\end{tabular}

\end{adjustbox}
\caption{\label{results-100}Results for {2,5,10,20,40,60,80,100} candidate responses with the original selection model}

\end{table}

\begin{table*}[h]
\centering
\begin{adjustbox}{width=\textwidth}
\begin{tabular}{rllll|llllll}
\cline{1-10}
\multicolumn{1}{l|}{}                                             & \multicolumn{9}{c}{50\% NOTA Test Results (\%)}                                            &  \\ \cline{1-10}
\multicolumn{1}{l|}{}                                             & R@10 & R@2 & N@10  & N@2   & N F1@10 & N F1@2 & G F1@10 & G F1@2 & Average F1 &  \\\cline{2-10}
\multicolumn{1}{l|}{Selection model trained with original data} &         &       &       &       &         &        &         &        &             &  \\
\multicolumn{1}{r|}{Direct Predict}                               & 56.12 & 66.77 & 61.48 & 78.00 & 52.82 & 80.22 & 67.46 & 75.21 & 68.93      &  \\
\multicolumn{1}{r|}{+Logistic Regression on Top of Logits}        & 55.98 & 66.72 & \textbf{87.81} & \textbf{88.19} & \textbf{86.96} & \textbf{87.26} & \textbf{88.56} & \textbf{88.99} & \textbf{87.94} &  \\
\multicolumn{1}{r|}{+Logistic Regression on Top of Softmax}       &50.94 & 51.93 & 74.30 & 74.33 & 74.46 & 74.38 & 74.15 & 74.29 & 74.32&  \\
\multicolumn{1}{r|}{Logits Threshold (=0.5)}                       & 50.10 & 55.72 & 64.28 & 73.25 & 62.84 & 76.73 & 65.61 & 68.56 & 68.43 & \\
\multicolumn{1}{r|}{+Logistic Regression on Top}                  & 62.81 & 77.70 & 80.45 & 79.95 & 80.49 & 79.92 & 80.42 & 79.99 & 80.20 &  \\
\multicolumn{1}{r|}{Softmax Threshold (=0.55)}                    & 48.76 & 48.76 & 60.10 & 70.67 & 59.69 & 75.63 & 60.50 & 63.17 & 64.74      &  \\
\multicolumn{1}{r|}{+Logistic Regression on Top}                  &\textbf{63.64} & \textbf{69.47} & 78.50 & 78.54 & 80.17 & 80.20 & 76.52 & 76.57 & 78.36 &  \\ \cline{1-10}

\multicolumn{1}{l|}{Selection model trained with data containing \emph{\_NOTA}}          &         &       &       &       &         &        &         &        &             &  \\
\multicolumn{1}{r|}{Direct Predict}                               & 55.43 & 65.03 & 63.07 & 78.37 & 54.28 & 80.91 & 69.03 & 75.04 & 69.81       &  \\
\multicolumn{1}{r|}{+Logistic Regression on Top of Logits}        & 40.66 & 47.90 & 78.19 & 77.45 & 78.80 & 78.02 & 77.53 & 76.85 & 77.80 &  \\
\multicolumn{1}{r|}{+Logistic Regression on Top of Softmax}       &51.63 & 53.90 & 77.94 & 78.00 & 78.21 & 78.15 & 77.67 & 77.85 & 77.97&  \\
\multicolumn{1}{r|}{Logits Threshold (=2.0)}                              & 48.44 & 55.99 & 61.32 & 71.31 & 57.75 & 74.35 & 64.32 & 67.46 & 65.97       &  \\
\multicolumn{1}{r|}{+Logistic Regression on Top}                  &60.73 & \textbf{76.12} & \textbf{79.22} & \textbf{78.03} & 79.11 & 77.85 & 79.33 & \textbf{78.21} & \textbf{78.62}&  \\
\multicolumn{1}{r|}{Softmax Thtrshold (=0.5)}                            & 48.18 & 48.18 & 59.06 & 70.16 & 57.32 & 75.19 & 60.67 & 62.56 & 63.94      &  \\
\multicolumn{1}{r|}{+Logistic Regression on Top}                  & \textbf{61.08} & 68.45 & 78.01 & 78.00 & \textbf{79.75} & \textbf{79.74} & 75.94 & 75.93 & 77.84 &  \\\cline{1-10}

\multicolumn{1}{l|}{Pairwise Model}          &         &       &       &       &         &        &         &        &             &  \\
\multicolumn{1}{r|}{Direct Predict}                               & 35.73 & 40.91 & 61.72 & 68.25 & 63.54 & 75.07 & 59.72 & 56.30 & 63.66       &  \\
\multicolumn{1}{r|}{+LogReg on Top of Logits}        &35.64 & 40.73 & \textbf{94.08} & \textbf{94.14} & \textbf{93.72} & \textbf{93.79} & \textbf{94.40} & \textbf{94.46} & \textbf{94.09} &  \\
\multicolumn{1}{r|}{+LogReg on Top of Softmax}       &25.42 & 27.14 & 85.06 & 85.02 & 85.41 & 85.34 & 84.69 & 84.67 & 85.03&  \\
\multicolumn{1}{r|}{Logits Threshold (=1.0)}                              & 41.64 & 48.57 & 61.50 & 70.01 & 57.77 & 74.36 & 64.62 & 63.88 & 65.16     &  \\
\multicolumn{1}{r|}{+LogReg on Top}                  & 51.58 & \textbf{73.33} & 77.15 & 77.27 & 76.74 & 76.88 & 77.55 & 77.64 & 77.20&  \\
\multicolumn{1}{r|}{Softmax Threshold (=0.4)}                            & 39.70 & 40.05 & 54.96 & 65.90 & 51.83 & 72.30 & 57.70 & 55.66 & 59.37      &  \\
\multicolumn{1}{r|}{+LogReg on Top}                  &\textbf{52.00} & 63.79 & 74.40 & 74.33 & 76.43 & 76.41 & 71.99 & 71.85 & 74.17&  \\ \cline{1-10}

\multicolumn{1}{l|}{Dropout Model}          &         &       &       &       &         &        &         &        &             &  \\
\multicolumn{1}{r|}{Direct Predict}                               & \textbf{28.57} & \textbf{93.47} & 50.13 & 62.42 & 1.48 & 45.50 & 66.61 & 71.32 & 46.23      &  \\
\multicolumn{1}{r|}{+LogReg on Top of Logits}        & 19.21 & 77.20 & \textbf{66.89} & \textbf{66.72} & \textbf{61.87} & 61.59 &\textbf{70.74} & \textbf{70.65} & \textbf{66.21}  \\
\multicolumn{1}{r|}{+LogReg on Top of Softmax}       & 21.73 & 29.37 & 50.49 & 54.83 & 56.37 & 63.73 & 42.79 & 40.15 & 50.76  \\
\multicolumn{1}{r|}{Logits Variance Threshold (=0.1)}                              & 13.73  & 22.11 & 51.89 & 50.27 & 57.15  & 59.13  & 45.15   & 36.51  & 49.48       &  \\
\multicolumn{1}{r|}{+LogReg on Top} & 20.87 & 60.78 & 56.13 & 55.86 & 40.18 & 39.29 & 65.37 & 65.32 & 52.54 \\
\multicolumn{1}{r|}{Softmax Variance Threshold (=0.001)}  &   22.22      &      36.75 &     50.03  &54.56       &  38.98 & 57.64       &    57.69&  50.99      &    51.32         &  \\
\multicolumn{1}{r|}{+LogReg on Top}  & 23.84 & 26.07 & 57.21 & 56.79 & 60.87 & \textbf{66.47} & 52.81 & 39.23 & 54.85 &  \\
\end{tabular}

\end{adjustbox}
\caption{\label{complete-results-table}$@10$ and $@2$ represent metrics on 10 and 2 candidates respectively. $R$ represents recall, $N$ represents binary NOTA classification accuracy, $N F1$ represents the F1 score on the NOTA class, and $G F1$ represents the F1 score on the ground-truth-present class. Average F1 is obtained on the 4 F1 scores.}

\end{table*}

Table \ref{results-100} shows the original selection model's performance on different sizes of candidate response sets. The direct predict model is run as it does not need further tuning. Threshold approach, especially with softmax probability as threshold, will need separate rounds of tuning on the threshold. Table \ref{complete-results-table} shows the complete results for all models on the test set, both for 2 candidates and for 10 candidates. Here, the average F1 is averaged on all 4 F1 scores. For each model architecture, the best performing setting for each metric is in bold.

\clearpage

\end{document}